%% file: unnatural-language.tex
\documentclass{article}

\usepackage[accepted]{icml2019}

\usepackage[utf8]{inputenc} % allow utf-8 input
\usepackage[T1]{fontenc}    % use 8-bit T1 fonts
\usepackage{hyperref}       % hyperlinks
\usepackage{url}            % simple URL typesetting
\usepackage{booktabs}       % professional-quality tables
\usepackage{tabularx}
\usepackage{amsfonts}       % blackboard math symbols
\usepackage{nicefrac}       % compact symbols for 1/2, etc.
\usepackage{microtype}      % microtypography
\usepackage{graphics}
\usepackage{xspace}
\usepackage{amsmath}
\usepackage{amssymb}
\usepackage{amsthm}
\usepackage{graphicx}
\usepackage{fancyvrb}
\usepackage{fvextra}
\usepackage{inconsolata}
\usepackage{wrapfig}

\newcommand\bert{\textsc{bert}\xspace}

\newcommand\model{f}
\newcommand\inputs{\mathcal{X}}
\newcommand\outputs{\mathcal{Y}}
\newcommand\fakeinputs{\widetilde{\mathcal{X}}}
\newcommand\fakedata{\widetilde{\mathcal{D}}}
\newcommand\realdata{\mathcal{D}}
\newcommand\fakeinput{\tilde{x}}
\newcommand\fakeoutput{\tilde{y}}
\newcommand\realinput{x}

\newcommand\eg{e.g.,\ }

\newcommand\embed{\texttt{embed}\xspace}

\DeclareMathOperator*{\argmin}{arg\,min}
\usepackage{xspace}

\newlength\myindent
\usepackage{xcolor}

\newcommand{\overnight}{\textsc{Overnight}\xspace}
\newcommand{\babyai}{\textsc{BabyAI}\xspace}

\icmltitlerunning{Unnatural Language Processing}

\nocite{*}
\begin{document}

\twocolumn[
\icmltitle{Unnatural Language Processing: \\Bridging the Gap Between Synthetic and Natural Language Data}
\begin{icmlauthorlist}
\icmlauthor{Alana Marzoev}{mit}
\icmlauthor{Samuel Madden}{mit}
\icmlauthor{M. Frans Kaashoek}{mit}
\icmlauthor{Michael Cafarella}{michigan}
\icmlauthor{Jacob Andreas}{mit}
\end{icmlauthorlist}
\icmlaffiliation{mit}{Massachusetts Institute of Technology}
\icmlaffiliation{michigan}{University of Michigan}
\icmlcorrespondingauthor{Alana Marzoev}{marzoev@mit.edu}
%\icmlauthor{Sam Madden}{madden@csail.mit.edu}
%\icmlauthor{M. Frans Kaashoek}{kaashoek@mit.edu}
%\icmlauthor{Michael Cafarella}{michjc@michigan.edu}
%\icmlauthor{Jacob Andreas}{jda@mit.edu}
\vskip 0.3in
]

\printAffiliationsAndNotice{}

\begin{abstract} 
   Large, human-annotated datasets are central to the development of natural language processing models. Collecting these datasets can be the most challenging part of the development process. We address this problem by introducing a general-purpose technique for ``simulation-to-real" transfer in language understanding problems with a delimited set of target behaviors, making it possible to develop models that can interpret natural utterances without natural training data. 
    
   We begin with a synthetic data generation procedure, and train a model that can accurately interpret utterances produced by the data generator. To generalize to natural utterances, we automatically find \textit{projections} of natural language utterances onto the support of the synthetic language, using learned sentence embeddings to define a distance metric. 
   With only synthetic training data, our approach matches or outperforms state-of-the-art models trained on natural language data in several domains. These results suggest that simulation-to-real transfer is a practical framework for developing NLP applications, and that improved models for transfer might provide wide-ranging improvements in downstream tasks.

\end{abstract}

\input{intro}

\section{Sim-to-real transfer for NLP}
\label{section:approach}

Using data from a synthetic grammar as described in \autoref{sec:grammars}, we can train a model such that for any desired output, there is \emph{some} input sentence that produces that prediction. The behavior of this model will be undetermined on most inputs. To interpret an out-of-scope utterance $x$ using a model trained on synthetic data, it is sufficient to find some synthetic $\tilde{x}$ with the same meaning as $x$, and ensure that the model's prediction is the same for $x$ and $\tilde{x}$.
In other words, all that is required for sim-to-real transfer is a model of \emph{paraphrase} relations and a synthetic data distribution rich enough to contain a paraphrase of every task-relevant input.

In this framework, one possible approach to the sim-to-real problem would involve building a \emph{paraphrase generation model} that could generate natural paraphrases of synthetic sentence, then training the language understanding model on a dataset augmented with paraphrases (\citeauthor{Basik18}, \citeyear{Basik18}; \citeauthor{YuSu17}, \citeyear{YuSu17}). However, collecting training data for such a paraphrase model might be nearly
as challenging as collecting task-specific annotations directly. Instead, we propose to find \emph{synthetic} paraphrases of \emph{natural} sentences, a process which requires only a model of sentence similarity and no supervised
paraphrase data at all.

Formally, we wish to produce a \textbf{wide-coverage model} $\model : \inputs \to \outputs$, where $\inputs$ is the 
space of of natural language inputs (\eg questions or instructions) and $\outputs$ is the
space of outputs (\eg meaning representations, action sequences, or dialogue responses). 
We accomplish this by defining a space of synthetic sentences, $\fakeinputs{}$.
and train a \textbf{synthetic model} $\tilde{f} : \fakeinputs{} \to \outputs$ on an arbitrarily large dataset of synthetic training examples 
$\fakedata = \{(\fakeinput_i, \fakeoutput_i)\}$.
To generalize to real sentences, all that is necessary is a function $\pi : \mathcal{X} \to \fakeinputs{}$ that ``projects'' real sentences onto their synthetic paraphrases.
We then define:
\begin{equation} 
    \label{eq:basic}
    \model(x) = \tilde{f}(\pi(x)) \ .
\end{equation}

The synthetic model $\tilde{f}$ can be trained using standard machine learning techniques as appropriate for the task. It remains only to define the projection function $\pi$.
Inspired by recent advances in language modeling and representation learning, we choose to use pretrained sentence representations as the basis for this projection function, mapping from natural language to synthetic utterances based on similarity in embedding space, under the assumption that rich contextual representations of language can be used to cope with distributional differences between natural and synthetic language.

In the remainder of section, we describe the steps needed to construct a working implementation of the projection function $\pi$ for language understanding problems of increasing complexity. Concretely, in \autoref{sec:para}, we introduce a reformulation of the problem of finding paraphrases as one of similarity search in embedding space. In \autoref{sec:amortized}, we reduce the computational complexity of the proposed similarity search, amortizing much of the cost into a single $O(|\fakeinputs|)$ preprocessing step. Then, in \autoref{sec:fast}, we introduce hierarchical projections, which facilitate the use of this approach in domains with complex grammars, in which even a single enumeration of $\fakeinputs$ is intractable. Finally, \autoref{sec:matching} describes an improvement to our model that helps to capture attribute-value distinctions that are nontrivial to extract from top-level sentence embeddings alone. 

\subsection{Paraphrase via similarity search}
\label{sec:para}
\begin{wrapfigure}{l}{0.25\columnwidth}
\vspace{-1em}
\includegraphics[width=0.25\columnwidth,clip,trim=.in 7.45in 10.2in .15in]{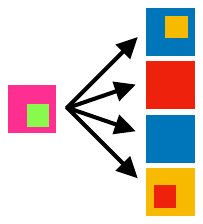}
\vspace{-2em}
\end{wrapfigure}

Many self-supervised pretraining schemes for NLP produce vector representations of sentences in which distance in vector space closely tracks semantic similarity (see e.g.\ recent results on the semantic textual similarity benchmark; \citeauthor{agirre2016semeval}, \citeyear{agirre2016semeval}).
We propose to project from the natural data distribution $\realdata$ onto the set of synthetic sentences $\fakeinputs$ with respect to a distance function $\delta$ defined by a pretrained sentence embedding model $\embed : \inputs \to \mathbb{R}^d$. That is, given a natural language input $\realinput$, we define:
\begin{equation}
\label{eq:projection}
    \pi_\textrm{full}(x) = \argmin_{\fakeinput \in \fakeinputs} ~ \delta(\embed(\realinput), \embed(\fakeinput))
\end{equation} and finally predict
$\model(\realinput) = \tilde{\model}(\pi_\textrm{full}(x))$
as in \autoref{eq:basic}. This framework is agnostic to choice of embedding model and distance metric. For all experiments in this paper, \texttt{embed} returns the average of contextual word representations produced by the \texttt{bert-base-uncased} model \cite{Devlin18BERT}, and $\delta(u, v)$ is the cosine distance $1 - u^\top v / (\|u\| \|v\|)$ (or the variant described in \autoref{sec:matching}).

This projection method is straightforward but computationally expensive: each projection requires enumerating and embedding every utterance that can be generated by the synthetic grammar. Depending on the problem domain, the size of $\fakeinputs$ can vary significantly. For simpler problems, there may be hundreds to thousands of examples. For complex problems; $|\fakeinputs|$ might be larger or even infinite, making explicit enumeration intractable or impossible.

\subsection{Amortized inference with locality sensitive hashing}
\label{sec:amortized}
\begin{wrapfigure}{l}{0.25\columnwidth}
\vspace{-1em}
\includegraphics[width=0.25\columnwidth,clip,trim=.in 7.45in 10.2in .15in]{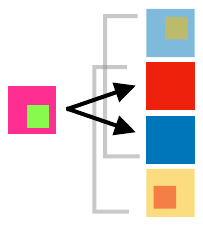}
\vspace{-2em}
\end{wrapfigure}

To reduce the $O(n)$ search cost of the $\argmin$ in \autoref{eq:projection}, we use \emph{locality sensitive hashing} (LSH; \citeauthor{gionis1999similarity}, \citeyear{gionis1999similarity}) to reduce the search space.
Rather than requiring a search across every candidate synthetic sentence for each new natural language input, locality sensitive hashing allows for a one-time $O(|\fakeinputs|)$ preprocessing step to compute hashes for synthetic sentences, such that sentences with similar embeddings fall into nearby buckets.
We then need search over only a constant number of nearby buckets of any given input natural language sentence to find all candidate synthetic sentences.

To implement locality sensitive hashing we use Simhash~\citep{moses_similarity_2002}, an LSH technique based on random projections of vectors.
Simhash takes as input a high dimensional vector in $\mathbb{R}^d$ and outputs an $f$-dimensional vectors of bits, called a \emph{fingerprint}, with the property that similar input vectors with regards to cosine similarity generate similar fingerprints.
To accomplish this, Simhash generates $f$ random hyperplanes in $d$-dimensional space (denoted $\ell_1 \cdots \ell_f$), computes the dot product of each hyperplane with the input vector, and outputs an $f$-bit hash corresponding to the sign of each dot product:
\begin{equation}
    h(x) = [ \textrm{sgn}(\ell_1^\top \texttt{embed}(x)),~\cdots,~\textrm{sgn}(\ell_f^\top \texttt{embed}(x)) ]
\end{equation}
Since nearby points tend to lie on the same side of random hyperplanes, the probability that two points have the same fingerprint is proportional to the angle between them, and thus directly related to their cosine similarity \citep{moses_similarity_2002}.
By bucketing datapoints according to their fingerprints, the exhaustive search in \autoref{eq:projection} can be restricted to those points with the same signature as $x$:
\begin{equation}
    \pi_\textrm{lsh}(x) = \argmin_{\fakeinput \in \fakeinputs:~h(\fakeinput) = h(x)} \delta(\embed(x), \embed(\fakeinput))
\end{equation}
Constructing a data structure to support LSH still requires exhaustively enumerating the full set of candidate synthetic utterances once, but reduces each subsequent lookup to time proportional to bucket size (controlled by $f$), which can be tuned to trade off between speed and recall.
% (which can be made arbitrarily small, in exchange for decreased accuracy, by increasing the fingerprint size).

% Simhash takes as input a high dimensional $d$-dimensional vector and outputs an $f$-dimensional vectors of bits, called a \emph{fingerprint}, with the property that similar input vectors generate similar fingerprints.
% To accomplish this, Simhash generates $f$ random hyperplanes in $d$-dimensional space, computes the dot product of each hyperplane with the input vector, and outputs $f$ bits corresponding to whether each dot product is positive or negative, exploiting the fact that similar points tend to be on the same side of random hyperplanes.
% With each input vector compressed to a small binary fingerprint, it is then easy to search for nearby points in high dimensional space by considering vectors with identical or similar fingerprints to a given input vector.

\subsection{Fast inference with hierarchical projection}
\label{sec:fast}
\begin{wrapfigure}{l}{0.33\columnwidth}
\vspace{-1em}
\includegraphics[width=0.33\columnwidth,clip,trim=.in 7.85in 10.05in .15in]{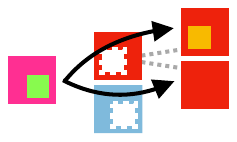}
\vspace{-1em}
\end{wrapfigure}

To further optimize the search process of the $\argmin$ within $\pi$, and to extend the projection techniques to domains where even a single enumeration of $\fakeinputs$ is intractable, we introduce
a \emph{hierarchical} procedure for computing $\pi$.
Rather than construct the LSH search data structure ahead of time, we construct a subset of it dynamically for each input query.
This procedure is not possible for general nearest neighbor search problems, but here we can rely on special structure: synthetic utterances are generated from a grammar, and the \embed function can be trained to provide a meaningful measure of similarity \emph{even for incomplete derivations under the grammar}. 

To perform hierarchical search, we assume that our synthetic data is generated from a context-free grammar \cite{hopcroft2001introduction}.
For example, the example sentence \emph{pick up the red ball} may be generated by the sequence of derivation steps \texttt{\$root} $\to$ \emph{pick up the \texttt{\$item}} $\to$ \emph{pick up the ball} (arrows $\to$ denote reachability in one step and \texttt{\$dollar} signs denote nonterminal symbols).

We use this derivation process to compute the $\argmin$ from previous equations iteratively, by repeatedly selecting a nonterminal to instantiate, instantiating it, and repeating the process until a complete synthetic sentence is generated. We illustrate our approach through an example:

Given an input natural language utterance, \emph{I want the dark blue crate adjacent to the opening} (\autoref{fig:my_label}), we first rank expansions of the root symbol of the synthetic grammar. 
Here we rely specifically on the use of a masked language model for sentence representations: we replace each nonterminal symbol with a \texttt{[MASK]} token to obtain meaningful similarity between complete real sentences and partially instantiated synthetic ones.
 
\begin{figure}[b]
    \centering
    \vspace{-.5em}
    \includegraphics[width=.85\columnwidth,trim=.2in 4.5in 4.8in .2in]{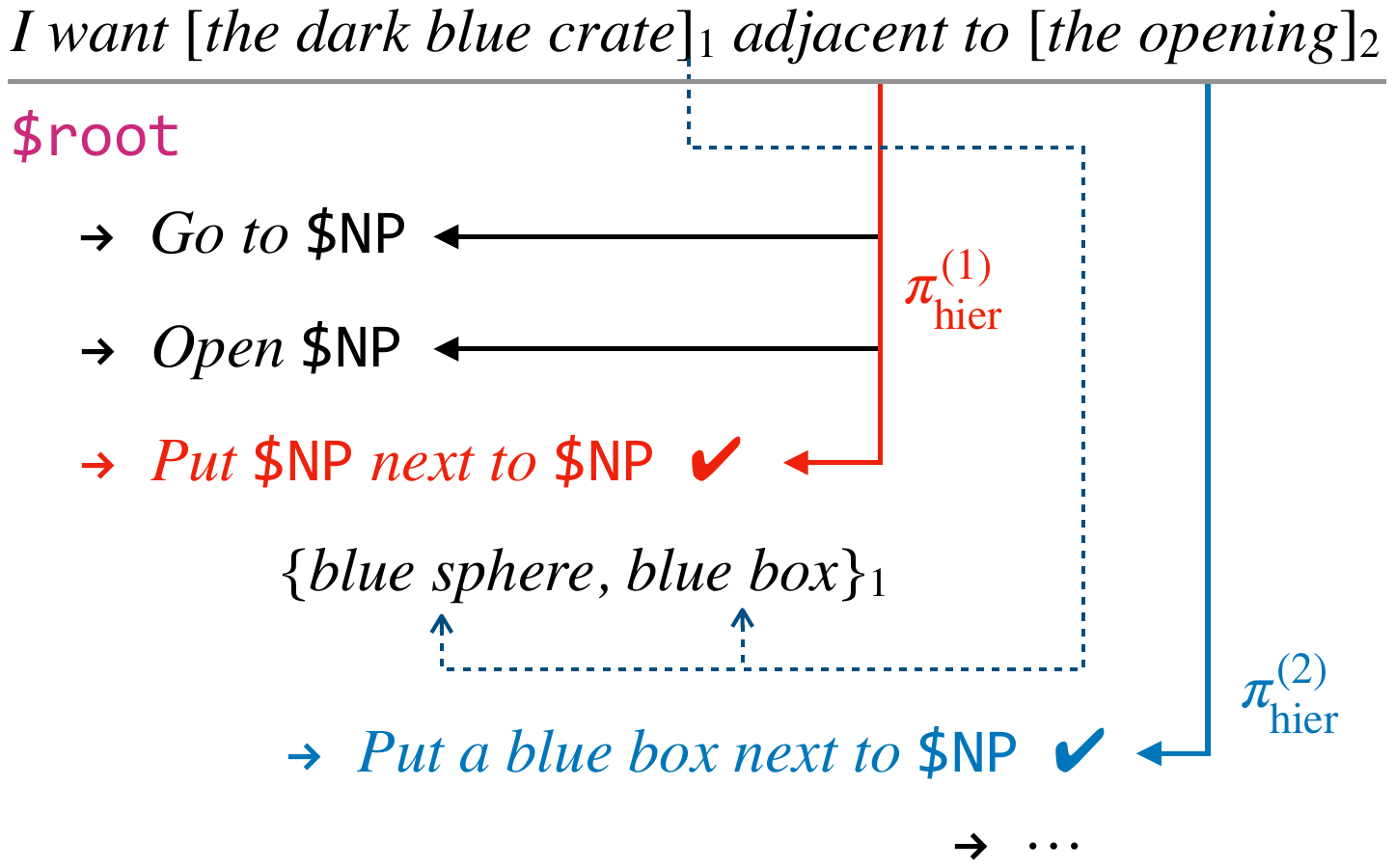}
    \vspace{-.5em}
    \caption{Hierarchical projection. Given a natural language input, we search for a high-scoring utterance generated by a fixed CFG, using similarity between sentences and partial derivations as a search heuristic. An additional heuristic scores noun phrases locally by measuring their similarity with noun chunks extracted from the input sentence.}
    \label{fig:approach-ex}
    \vspace{-1em}
\end{figure}
 
Letting $\fakeinputs(\texttt{\$root})$ denote the set of (complete or incomplete) sentences that can be derived from the incomplete derivation \texttt{\$root} in a single step, we have:
 \begin{equation}
     \pi_\textrm{hier}^{(1)}(x) = \argmin_{\fakeinput \in \fakedata(\texttt{\$root})} \delta(\embed(x), \embed(\fakeinput))
     \label{eq:hier}
 \end{equation}
 as before. For the example in \autoref{fig:approach-ex}, $\pi_\textrm{hier}^{(1)}(x) =$ \textit{put the \texttt{\$item} next to \texttt{\$item}}.
 
The process is then repeated for all unexpanded nonterminals in the partial derivation until a
complete sentence is generated. 
This procedure corresponds to greedy search over sentences generated by the CFG, with
measured similarity between partial derivations and the target string $x$ as a search heuristic.
This procedure can be straightforwardly extended to a beam search by computing the $k$ lowest-scoring
expansions rather than just the $\argmin$ at each step.

If the grammar has a high branching factor (i.e.\ the number of expansions for each nonterminal symbol is large), it may 
also be useful to incorporate other search heuristics. For all experiments in this paper, we restrict the set of expansions
for noun phrase nonterminals.
We begin by using a pretrained chunker \cite{sang2000introduction} to label noun phrases in the natural language input $\realinput$ (in \autoref{fig:approach-ex} running example this gives \emph{the dark blue crate} and \emph{the opening}).
When expanding noun phrase nonterminals, we first align the nonterminal to a noun chunk in $\realinput$ based on
similarity between the chunk and all right-hand sides for the nonterminal, and finally select the single right-hand
side that is most similar to the aligned noun chunk. Greedy population of noun phrases ensures greater beam diversity of sentence-level templates in beam hypotheses.
We additionally discard any hypotheses in which there is a mismatch between the number of noun phrases in the query sentence $x$ and the number of groups of adjacent nonterminals in the top-level partial derivation $\fakedata^{(1)}$.  For example, if the template calls for two distinct entities, but the natural language utterance only contains one (e.g. \emph{go to the green door}), then the derivation is immediately discarded.

Given a grammar of size $g$, derivations of size $d$, and SimHash bucket size $b$, hierarchical projections reduce inference cost from $O(|\fakedata|)$ to $O(bd < gd)$. 

\subsection{Tunable models with matching scores}
\label{sec:matching}

\begin{wrapfigure}{l}{0.33\columnwidth}
\vspace{-1em}
\includegraphics[width=0.33\columnwidth,clip,trim=.in 8in 10.15in .15in]{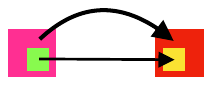}
\vspace{-2em}
\end{wrapfigure}

Hierarchical representations of synthetic utterances makes it possible to improve scoring as well as search.
Consider the set of candidates:
\begin{enumerate}
    \item \emph{go through the yellow door and pick up the red ball}
    \item \emph{go through the red door and pick up the yellow ball}
\end{enumerate}
and a natural language input referencing a \emph{red ball} and {yellow door}. Candidate (1) should be prioritized over candidate (2), but in early experiments we found that fine-grained attribute--value distinctions are not always captured by top-level sentence embedding similarities. To improve this, we also experiment with a modified version of the similarity measure $\delta$ that incorporates fine-grained lexical similarity in addition to sentence similarity. 

To compute this new distance function $\delta'$, we model the problem as one in finding a minimum weight matching in bipartite graphs. 
Suppose we have a set of (complete) candidate synthetic utterances derived from either a flat or hierarchical projection procedure as defined above.
As above, we extract noun chunks from the natural language query and each synthetic utterance, representing each chunk as a node in a graph connected to each node from the utterance with edge weights equivalent to the distance between the respective \emph{phrases} in embedding space. We use the Hungarian algorithm~\citep{kuhn_hungarian_1955} to solve this matching problem, and incorporate the computed minimum syntactic cost $\delta_{sub}$ from each candidate sentence into the overall distance function $\delta$. In particular, we redefine the similarity function as
\begin{align}
\delta'(x, \fakeinput) = &\delta(\embed(x), \embed(\fakeinput)) \nonumber \\& + \alpha \sum_{(x', \tilde{x}') \in M} \delta(\embed(x'), \embed(\tilde{x}')
\end{align}
where $M$ is the matching found by the Hungarian algorithm and $\alpha$ is a hyperparameter that can be tuned as required based on the grammar of the problem domain. 

This scoring strategy can have detrimental effect if  $\alpha$ is set too high --- one example of a where $\delta_{sub}$ incorrectly overpowers $\delta$ is when ``go through the yellow door and pick up the red ball” is projected onto an utterance containing those extra details that is not semantically equivalent, such as ``put the red ball next to the yellow door''. However, in early experiments on \babyai dataset we found the additional fine-grained control provided by matching scores led to slight improvements in projection accuracy: see \autoref{sec:eval} for experiments.

\input{evaluation}

\input{related}

\section{Discussion}
We have described a family of techniques for sim-to-real transfer in natural language processing contexts, a comparative analysis of these techniques, a new benchmark for measuring sim-to-real transfer performance, and a practical software framework for bootstrapping synthetic grammars.

The projection procedure described in this paper suggests a new approach to bootstrapping natural language applications. In this paradigm, key insights about the relationship between the world and language are explicitly encoded into the declarative synthetic data generation procedure rather than implicitly in the model's structure or through the use of a human-annotated dataset, and can take advantage of advances in machine learning and structured knowledge about human language. Developers who want to build applications in new domains can therefore hand-engineer synthetic grammars, use the generated data to train domain-specific machine learning models, and use projections to paraphrase test time natural language examples into their synthetic counterparts. This makes it possible to recover substantial amounts of model accuracy that otherwise would have been lost when switching from the distribution of synthetic to natural language utterances.

\bibliography{bib,jacob}
\bibliographystyle{icml2019}
\end{document}

%% file: intro.tex
\section{Introduction}
Data collection remains a major obstacle to the development of learned models 
for new language processing applications. Large text corpora are available 
for learning tasks like language modeling and machine translation \cite{CCB11WMT,Chelba19BillionWord}. But other classes of NLP models---especially those that interact with the outside world, whether via API calls 
(e.g.,\ for question answering) or physical actuators (e.g.,\ for personal robotics)---require custom datasets that capture both the full scope of desired behavior in addition to the possible variation in human language. Collecting these large, human-annotated training sets can be an expensive and time-consuming undertaking \cite{Zelle95GeoQuery}. 
In domains governed by well-defined mathematical models, such as physics simulators and graphics engines, one solution to the data scarcity problem problem is ``simulation-to-real" transfer \cite{Tzeng2015Sim2Real}. In sim-to-real approaches, knowledge gained in a simulated environment is later applied in the real world with the ultimate aim of generalizing despite discrepancies between the simulated environment and reality.

\begin{figure*}[t!]
    \centering
    \vspace{-.5em}
    \includegraphics[width=0.9\textwidth,clip,trim=0in 4.3in 1.3in 0in]{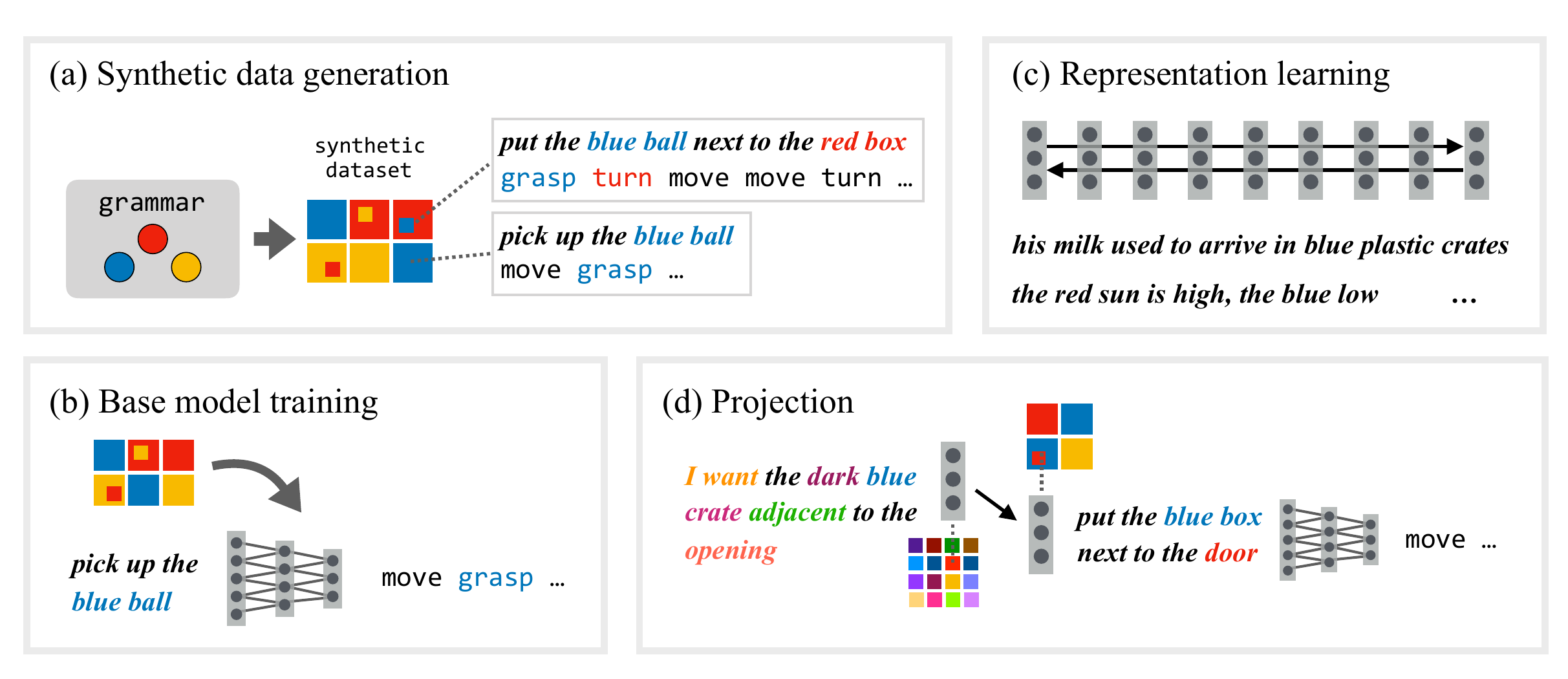}
    \vspace{-1em}
    \caption{Overview of our approach to synthetic-to-real transfer in language understanding tasks. (a) Training examples are generated from a synthetic data generation procedure that covers desired model behaviors but a limited range of linguistic variation. (b) This data is used to train a model that can correctly interpret synthetic utterances. (c) Separately, sentence representations are learned using a masked language modeling scheme like \bert. (d) To interpret human-generated model inputs from a broader distribution, we first \emph{project} onto the set of sentences reachable by the synthetic data generation procedure, and then interpret the projected sentence with the trained model.}
    \label{fig:my_label}
\end{figure*}

In this paper, we explore sim-to-real transfer for natural language processing. We use simple, high-precision grammars as ``simulators" to generate synthetic training data for question answering and instruction following problems.
While synthetic data 
generation provides potentially unlimited supervision for the learning
of these behaviors,
interpretation of synthetic utterances may itself constitute a
challenging machine learning problem when the desired outputs require 
nontrivial inference for parsing, planning or perception \cite{Luketina2019Survey}.
Given a model with high accuracy on the synthetic training distribution, we interpret natural user utterances from outside
this distribution by mapping each natural utterance to a synthetic one and interpreting the synthetic utterance with the learned model. Using pretrained sentence embeddings \cite{Devlin18BERT}, we define an (approximately) meaning-preserving projection operation from the set of all sentences to those the model has been trained to interpret.
Together, labeled synthetic utterances and unsupervised representation learning enable generalization to real language.

Through experiments, we demonstrate the effectiveness of sim-to-real transfer on a variety of domains. On a suite of eight semantic parsing datasets \cite{Wang15Overnight}, sim-to-real matches the performance of a \emph{supervised} semantic parser on three of eight tasks using no 
natural training data. On a 
grounded instruction following benchmark involving challenging navigation in
a gridworld environment \cite{Chevalier2018BabyAI}, our approach to sim-to-real transfer again surpasses the performance of a standard model fine-tuned with human annotations.

These results indicate the promise of sim-to-real as a development paradigm for natural language processing applications. 
By leveraging two cheap, readily-available sources of supervision---unaligned 
natural language text and synthetic language about target tasks---it is
possible to build broad-coverage systems for language understanding without a labor-intensive annotation process. 
Improved models for sim-to-real could lead to positive and wide-ranging effects on downstream tasks and applications. To encourage progress on models for transfer and to further reduce the developer effort associated with building new grounded language understanding, we release (1) a set of new human annotations for a popular policy learning benchmark with synthetic instructions and (2) code implementing the sentence-projection operation.\footnote{Available at https://github.com/unnatural-language/sim2real.}
 
\section{Grammar engineering for grounded language learning}
\label{sec:grammars}

Sim-to-real transfer requires to a simulator: an automated procedure that can generate labeled input--output examples.
The experiments in this paper focus on two \emph{language understanding} problems: question answering and instruction following. Both problems require mapping natural language inputs (e.g., \emph{go to the red door}) to concrete interpretations (e.g., a program $\texttt{(go-to (location red\_door))}$, which may be executed to produce a sequence of situated low-level actions $(\texttt{move, turn\_r, move, move})$). Many different sentences may be associated with the same program (\emph{find a red door}, \emph{navigate to the red door}, etc.).

A simulator for situated language understanding can be implemented as an injective expert-designed \emph{inverse} function, possibly multi-valued, mapping each program (e.g., $\texttt{(go-to (location yellow\_door))}$ to a set of distinct input sentences that induce the program (e.g., ``go to the location of the yellow door'').
This inverse function defines a \emph{synthetic data generation procedure} for the task, a direct mapping between canonical programs and plausible input sentences which can be used to generate training data.
Mature tools exist for designing such mappings, generally by adapting domain-general grammars \cite{bender2015proceedings}.
Synthetic grammars can be engineered to provide full coverage of the learner's \emph{output space}, generating a plausible input sentence for any target program. Closely related previous work \cite{Berant14Paraphrase} has shown that, even without full coverage of the \emph{input space}, such 
grammars can provide useful features for downstream learning.

While engineered grammars make it possible to generate large amounts labeled training data, they can provide at best limited coverage of natural language \cite{erbach1990grammar}: each program is mapped to a small number of plausible input sentences (as defined by the expert-written inverse function), meaning that the full spectrum of linguistic variation is not observed in the generated data.
This leads to catastrophic distributional shifts for models trained exclusively on this generated synthetic data: for example, on a drone instruction following benchmark, Blukis et al.\ report a test-time accuracy gap of over 54\% between models trained on synthetic data and those trained on real user utterances~(\citeyear{Blukis18Drone}).
In the next section, we describe a simple modeling approach that allows system-builders to use synthetic data to train models that are robust to natural inputs.

%% file: evaluation.tex
\section{Evaluation}
\label{sec:eval}

We evaluate our approach on two tasks with markedly different output spaces and data requirements: the \overnight semantic parsing benchmark \cite{Wang15Overnight} and the \babyai instruction following benchmark \cite{Chevalier2018BabyAI}. These experiments aim (1) to test the flexibility of the general sim-to-real framework across problems with complex compositionality and complex interaction, and (2) to measure the effectiveness of projection-based sentence interpretation relative to other ways of learning from pretrained representations and synthetic data. 

Both tasks come with predefined synthetic data generators intended for uses other than our projection procedure. The effectiveness of the approach described in this paper inevitably depends on the quality of the data synthesizer; by demonstrating good performance using off-the-shelf synthesizers from previous work, we hope to demonstrate the generality and robustness of the proposed approach. We expect that better
results could be obtained using a grammar optimized for peformance; we leave this for future work.

\subsection{Models}

Experiments on both benchmarks use the following models:

\paragraph{seq2seq} A baseline neural sequence model. For the semantic parsing task, this is a standard LSTM encoder--decoder architecture with attention \cite{Bahdanau14Attention}. (This model achieves comparable average performance to the dependency-based semantic parsing approach of \citeauthor{Wang15Overnight}.) We use a 256-dimensional embedding layer and a 1024-dimensional hidden state.  For the instruction following task, we reuse the agent implementation provided with the \babyai dataset, which jointly encodes instructions and environments states with a FiLM module \cite{Perez18Film}, and generates sequences of actions with an LSTM decoder. More details are provided in the original work of \citet{Chevalier2018BabyAI}. The seq2seq baseline measures the extent to which available synthetic data in each domain is already good enough to obtain sim-to-real transfer from standard models without specialization.

\paragraph{seq2seq+\bert} As discussed in \citet{Artetxe2019MultilingualBert}, the standard approach to the sort of zero-shot transfer investigated here is to train models of the kind described above not on raw sequences of input tokens, but rather on sequences of contextual embeddings provided by a pretrained representation model. The intuition is that these representations capture enough language-general information---e.g.\ about which words are similar and which syntactic distinctions are meaningful---that a learned model of the relationship between input representations and task predictions in the source (synthetic) domain will carry over to input representations in the target domain.

Concretely, seq2seq+\bert models are constructed for both tasks by taking the language encoder and replacing its learned word embedding layer with contextual embeddings provided by the \texttt{bert-base-uncased} model of \citeauthor{Devlin18BERT}. This is analogous to the strategy employed by \citet{Lindemann2019BertSemparse} and has been shown to be an effective way to incorporate pretraining into a variety of semantic prediction tasks.
Though the bulk of this paper focuses on the projection technique, the experiments we present are also, to the best of our knowledge, the first to systematically evaluate even this basic pretraining and transfer scheme in the context of synthetic data.

\paragraph{projection} The final model we evaluate is our projection approach described in \autoref{section:approach}. As discussed above, the projection model is built from the same pieces as the baselines: it relies on the base seq2seq model for interpreting synthetic data, the similarity function induced by \bert embeddings in the projection step. The set of synthetic sentences in the \overnight dataset is relatively small (a few hundred for the largest domains) so we apply the “flat projections” approach directly. \babyai, on the other hand, contains many more sentences (approximately $40$K), so we use hierarchical projections, evaluating the approach both with and without the linear sum assignment strategy applied. 

\begin{table*}
\footnotesize
\centering
\begin{tabularx}{\textwidth}{ll*{9}{X}}
\toprule
Data & Model & basketball & blocks & calendar & housing & pubs & recipes & restaurants & social & mean \\
\midrule
synth & projection &
0.47 &
\bf \underline{0.27} &
\bf 0.32 &
\bf \underline{0.36} &
\bf 0.34 &
\bf 0.49 &
\bf \underline{0.43} &
0.28 &
\bf 0.37 \\
 & seq2seq & 
0.27 &
0.08 &
0.22 &
0.16 &
0.24 &
0.21 &
0.23 &
0.07 &
0.19 \\
 & seq2seq+\bert &
\bf 0.60 & 
0.21 & 
0.31 & 
0.31 & 
\bf 0.34 & 
0.36 & 
0.36 & 
\bf 0.31 & 
0.35 \\
\midrule
real & seq2seq & 
0.45 &
0.10 &
0.27 &
0.19 &
0.30 &
0.38 &
0.25 &
0.20 &
0.27 \\
& seq2seq+\bert &
0.63 &
0.26 &
0.37 &
0.29 &
0.44 &
\underline{0.53} &
\underline{0.43} &
\underline{0.42} &
0.42 \\
\midrule
both & seq2seq & 
0.01 &
0.25 &
0.13 &
0.10 &
\underline{0.52} &
0.18 &
0.42 &
0.36 &
0.25 \\
& seq2seq+\bert &
\underline{0.67} &
0.23 &
\underline{0.43} &
0.29 &
0.48 &
0.29 &
0.36 &
0.38 &
0.39 \\
\bottomrule
\end{tabularx}
\caption{
\overnight semantic parsing logical form accuracies on human-generated questions for models trained with synthetic data, real data, or both. \textbf{Bold} values indicate the best-performing model under the \emph{synth} data condition, while \underline{underlined} values indicate the best-performing model under \emph{any} data condition. The projection approach consistently outperforms other models in the synthetic-only condition, and is comparable to a fully-supervised model in three domains.
}
\label{tab:overnight}
\end{table*}

\begin{table}
  \centering
  \footnotesize
  \begin{tabular}{llll}
  \toprule
  Data & Model & success & reward \\
  \midrule
  synth & projection ($\alpha = 0.1$) & \bf \underline{0.63}	& \bf \underline{0.72} \\
  & projection ($\alpha = 0.0$) & 0.59 & 0.70 \\
  & seq2seq & 0.56 & 0.67 \\
  & seq2seq+\bert  & 0.54 &	0.66 \\
  \midrule
  % both is 2048
  both & seq2seq & \bf 0.59	& \bf 0.71 \\
  & seq2seq+\bert & 0.53 & 0.65 \\
  \bottomrule
  \end{tabular}
  \caption{
    Instruction following accuracies on the \babyai \texttt{SynthLoc} task with human-generated instructions and models trained on synthetic or a mix of synthetic and real data. \textbf{Bold} values indicate the best-performing model for a given data condition and \underline{underlined} values indicate the best-performing model under any data condition. Projection ($\alpha = 0.1)$ gives the best results on both success (task completion) and discounted reward metrics, outperforming other models trained on only synthetic data as well as models fine-tuned on 2048 real examples.
  }
  \label{tab:babyai}
\end{table}

\subsection{Experiments}

\paragraph{Semantic parsing}

In semantic parsing, the goal is to learn a mapping from natural language utterances to formal, executable representations of meaning (\autoref{fig:overnight-ex}). These meaning representations can then be used to as inputs to database engines or formal reasoning systems. The \overnight benchmark consists of eight semantic parsing datasets covering a range of semantic phenomena, and is designed to test efficient learning from small numbers of examples---individual datasets range in size from 640--3535 examples.

\begin{figure}
\footnotesize
\begin{tabularx}{\columnwidth}{lX}
\toprule
real & \it show me the meeting starting latest in the day \\
synth & \it meeting that has the largest start time \\
LF & \tt    (max meeting (get start\_time)) \\
\midrule
real & what recipes posting date is at least the same as rice pudding \\
synth & recipe whose posting date is at least posting date of rice pudding \\
LF &   \tt  (filter recipe ($\geq$ \newline 
\strut~~(get posting-date) \newline 
\strut~~((get~posting-date)~RICE-PUDDING))) \\
\bottomrule
\end{tabularx}
\caption{Example sentences from the \emph{calendar} and \emph{recipe} \overnight domains. \emph{real} is a human-generated utterance, \emph{synth} is a synthetic utterance from the domain grammar of \emph{calendar}, and \emph{LF} is the target logical expression.}
\label{fig:overnight-ex}
\end{figure}

The semantic originally described by \citeauthor{Wang15Overnight} for this task was equipped with a ``canonical grammar''---a compact set of rules describing a mapping from logical forms to (somewhat stilted) natural language strings (\emph{meeting whose end time is smaller than start time of weekly standup}). In the original work, this grammar was used as a source of features for reranking logical forms. Here we instead use it as a source of \emph{synthetic training data}, enumerating strings from the grammar paired with logical forms, and using these pairs to train the base seq2seq model. The dataset also comes with a set of natural language annotations on train- and test-set logical forms.

Results are shown in \autoref{tab:overnight}. We report logical form exact match accuracies for the eight \overnight datasets.\footnote{The original work used a coarser-grained denotation match evaluation.} Three data conditions are evaluated: \emph{synth}, which uses synthetic-only training data, \emph{real}, which uses only human-annotated training data, and \emph{both}, which combines the two. Human annotations are used for the test evaluation in all data conditions. For the \emph{both} condition, we found that simply concatenating the two datasets was reliably better than any fine-tuning scheme. 

Models in the \emph{real} and \emph{both} sections have access to significantly richer supervision than the projections approach, and are intended to characterize a best-case data condition. It can be seen that the projection-based approach to learning from synthetic data consistently outperforms sequence-to-sequence models with or without pretrained representations. In fact, projection is comparable to a \emph{supervised} approach (in the \emph{real} and \emph{both} data conditions) in three domains. These results demonstrate that projection is an effective mechanism for sim-to-real transfer in tasks with a challenging \emph{language understanding} component, enabling the use of synthetic strings to bootstrap a model that can perform fine-grained linguistic analysis of real ones.

\paragraph{Instruction following}
\begin{figure}
\footnotesize
    \centering
    \includegraphics{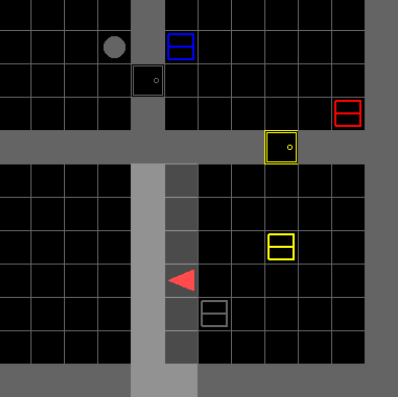} \\
    synth: \emph{put a yellow box next to a gray door} \\
    real: \emph{drive around and put the yellow block beside the top door}
    \caption{Example from the \babyai dataset. Agents are given tasks with language-like specifications, and must execute a sequence of low-level actions in the environment to complete them. We augment this dataset with a set of human instructions, and evaluate generalization of agents trained only on synthetic goal specifications to novel human requests.}
    \label{fig:babyai-ex}
\end{figure}

The \babyai instruction following dataset presents a range of challenging manipulation and navigation tasks in two-dimensional gridworld environments---agents are trained to achieve a wide variety of compositional goals (e.g. \emph{put a yellow box next to a gray door}, \autoref{fig:babyai-ex}) specified by sentences from an artificial grammar.

This dataset was originally designed to explore generalization in reinforcement and imitation learning, not natural language understanding, but it provides a flexible and extensible test-bed for studying questions around generalization in natural language as well. We collected a dataset of roughly 5000 natural language utterances by showing Mechanical Turk workers videos of agents executing plans in \babyai environments (specifically from the \texttt{SynthLoc} training set). Workers did not see the original \babyai commands, but were asked to generate new instructions that would cause a robot to accomplish the goal displayed in the video. The dataset of human annotations is released with this paper.

As in the \overnight experiments, we train (in this case via imitation learning) on either the underlying synthetic instructions alone, or a mix of synthetic and real instructions. We evaluate various models' ability to generalize to a human-annotated test set. 
Unlike the \overnight datasets, bootstrapping even a seq2seq model that achieves good performance on a test set with \emph{synthetic} instructions requires enormous amounts of data: the existing synthetic training set contains a million demonstrations. To adapt to human instructions, we fine-tune this baseline model (rather than concatenating training sets as above).

Results are shown in \autoref{tab:babyai}. Again, projection is the best way to incorporate pretrained representations: introducing \bert embeddings into the underlying sequence-to-sequence model makes performance worse. (We attribute this to overfitting to the limited set of word embeddings observed during training.) Again, as well, projection is effective enough to allow synthetic-only training to outperform fine-tuning on a small amount of human data. These results indicate that sim-to-real transfer is also an effective strategy for tasks whose sample complexity results from the difficulty of an underlying planning problem rather than language understanding as such.

%% file: related.tex
\section{Related work}

The technique described in this paper brings together three overlapping lines of work in machine learning and natural language processing:

\paragraph{Sim-to-real transfer in robotics}
The most immediate inspiration for the present work comes from work on simulation in robotics and other challenging control problems \cite{Tzeng2015Sim2Real}.
Simulators provide convenient environments for training agents because the poor sample efficiency of current policy learning algorithms are mitigated by faster-than-real-time interaction and unlimited data. ``Learning'' in simulation amounts to amortizing the inference problem associated with optimal control \cite{Levine18OptimalControl}: correct behavior is fully determined by the specification of the environment but poses a hard search problem.

However, simulators do not represent reality with complete accuracy, and agents trained to perform specific tasks in simulators often fail to perform these same tasks in the real world. Various techniques have been proposed for bridging this this ``reality gap" \cite{Tobin17Randomization}.
Language processing applications suffer from similar data availability issues. For grounded language learning in particular, ``end-to-end'' architectures make it hard to disentangle challenging \emph{language learning} problems from challenging \emph{inference} problems of the kind discussed above \cite{Andreas15Instructions}. Our approach aims to offload most learning of linguistic representations to pretraining without sacrificing the expressive power of models for either part of the problem.

\paragraph{Representation learning}

The last two years have seen remarkable success at learning general-purpose representations of language from proxy tasks like masked language modeling \cite{Peters18ELMO,Devlin18BERT}. For tasks with limited in-domain training data available, a widespread approach has been to begin with these pretrained representations, train on data from related source domains in which labels are available, and rely on similarity of pretrained representations in source and target domains to achieve transfer \cite{Artetxe2019MultilingualBert}. One surprising finding in the current work is that this strategy is of only limited effectiveness in the specific case of synthetic-to-real transfer: it is better to use these pretrained representations to find paraphrases from the target domain back to the synthetic source domain, and interpret sentences normally after paraphrasing, than it is to rely on transfer of representations themselves within a larger learned model.

\paragraph{Grammar engineering}

Finally, we view our work as offering a new perspective in a longstanding conversation around the role of grammar engineering and other rule-based approaches in NLP \cite{Chiticariu13RuleBasedIE}. Rule-based approaches are criticized for being impossible to scale to the full complexity of real language data, providing good coverage of ``standard'' phenomena but requiring prohibitive additional to model the numerous special cases and exceptions that characterize human language \cite{Norvig17Chomsky}.

However, engineered grammars and synthetic data generation procedures also offer a number of advantages. Mature engineering tools and syntactic resources are available for many major languages \cite{Flickinger11ERG}. Hand-written grammars are straightforward to implement and can serve as cost-effective tools for language generation in larger data-synthesis pipelines; they also enable developers to quickly incorporate new linguistic phenomena when they are found to be outside model capacity. Moreover, since the relative frequency syntactic and semantic phenomena can be specified by engineers, the poor accuracy on long-tail pheonomena that is observed in models trained on \emph{natural} datasets \cite{Bamman2017LongTail} can be mitigated by flattening out the relevant distributions in the data generation process.

The approach presented in this paper, and the prospect of more general sim-to-real transfer in NLP, offers to turn the partial solution offered by current grammar engineering approaches into a more complete one: model builders with limited data collection ability can construct high-quality synthetic data distributions with general-purpose language resources, project project onto these distributions with high accuracy, and eventually obtain good end-to-end performance without end-to-end data collection or training.